\documentclass{article} %
\usepackage{iclr2026_conference,times}

\usepackage{amsmath,amsfonts,bm}

\def\eqref#1{equation~\ref{#1}}

\def\1{\bm{1}}

\DeclareMathAlphabet{\mathsfit}{\encodingdefault}{\sfdefault}{m}{sl}
\SetMathAlphabet{\mathsfit}{bold}{\encodingdefault}{\sfdefault}{bx}{n}

\usepackage{hyperref}
\usepackage{url}
\usepackage{booktabs}
\usepackage{graphicx}
\usepackage{subcaption}
\usepackage[table]{xcolor}
\usepackage{wrapfig}
\usepackage{fontawesome5}
\usepackage{xspace}

\title{From Pixels to Patches: \faSwimmer~Pooling\\Strategies for
Earth Embeddings\thanks{No swimming required.}}

\iclrfinalcopy

\author{
  Isaac Corley$^\lambda$, Caleb Robinson$^\gamma$, Inbal Becker-Reshef$^\gamma$, Juan M. Lavista Ferres$^\gamma$\\
  $^\lambda$Wherobots, $^\gamma$Microsoft AI for Good Research Lab\\
  \\
}

\newcommand{\cellperf}[2]{\cellcolor{gray!\number\numexpr#1*3/4\relax}{#2}}

\begin{document}

\maketitle

\begin{abstract}
Geospatial foundation models increasingly expose pixel-level embedding products that can be downloaded and reused without access to the underlying encoder. In this setting, downstream tasks with patch- or region-level labels require a post-hoc aggregation step that maps dense pixel embeddings to a single representation. The default choice, mean pooling, discards within-patch variability and can underperform under spatial distribution shift. To study this setting, we introduce \emph{EuroSAT-Embed}: 81,000 embedding GeoTIFFs derived from three foundation models: AlphaEarth, OlmoEarth, and Tessera. Using these fixed embedding products, we benchmark 11 training-free pooling methods and 2 train-set-fitted baselines under both random and geographically disjoint test splits. Richer pooling schemes reduce the geographic generalization gap by over 50\% relative to mean pooling and improve accuracy by up to 6\% on spatial splits. We recommend a three-tier strategy: (1)~\textit{mean} as a baseline, (2)~\textit{stats} pooling (min/max/mean/std) as the default at 4$\times$ the embedding dimension, and (3)~\textit{covariance} pooling for peak accuracy. Across all three embedding products, simple distributional statistics improve spatial-split performance over mean pooling.
\end{abstract}

\section{Introduction}

\begin{wrapfigure}{r}{0.42\linewidth}
  \vspace{-12pt}
  \centering
  \includegraphics[width=\linewidth]{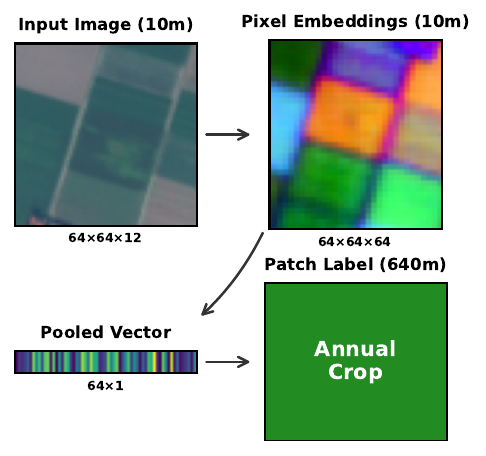}
  \caption{\textbf{Pixel-to-patch pooling.} The input-label resolution mismatch requires aggregating dense pixel embeddings (shown as PCA pseudo-RGB) to a lower resolution for downstream tasks.}\label{fig:hero}
  \vspace{-10pt}
\end{wrapfigure}

Geospatial foundation models (GFMs) turn satellite data time-series into embedding data products that can be reused across tasks and regions~\cite{fang2026earthembeddingsproductstaxonomy}.
Recent GFMs create dense satellite imagery embeddings at \emph{pixel} resolution, but many downstream tasks operate on objects or regions (fields, parcels, buildings, solar farms, gridded patches), requiring aggregation of pixel embeddings over polygons.
Pooling is therefore an inherent step when using GFM output for these tasks, but naive methods may erase relevant signals in the embeddings.

This problem compounds when labels exist at a coarser-than-pixel resolution, known as the \textit{input-label resolution mismatch}~\citep{workman2023handling}.
For example, EuroSAT~\citep{helber2019eurosat} contains 64$\times$64 patches of 10\,m pixels corresponding to a single land-cover label, which can be viewed as a \textit{zonal-statistics} aggregation operation~\citep{cressie2015statistics}.
Unlike raw spectral bands, pixel embeddings encode semantic features that vary within a land-cover patch.
This heterogeneity suggests that averaging can remove class signal that richer statistics keep.
This makes pooling a design decision for geospatial applications with coarse labels and/or temporal dimensions.

Pooling is underexplored for earth embeddings\footnote{A relevant exception is the recent OlmoEarth paper~\cite{herzog2025olmoearth} that reports searching over mean and max temporal pooling strategies as a hyperparameter in downstream tasks.}, though it is heavily studied in related domains.
In image retrieval, choices like generalized mean (GeM) pooling~\citep{radenovic2018gem} often matter as much as the learned encoder.
In audio and video processing, statistical pooling (concatenating means and standard deviations) effectively summarizes variable-length sequences~\citep{miech2017learnable}.
Mean pooling dominates for its simplicity, but its performance under spatial shift is not well measured.
Classical object-based image analysis~\citep{eo4geoobia} methods like Bag of Visual Words (BoVW)~\citep{csurka2004visual} aggregate hand-crafted features based on region statistics for vision tasks, but there is little work on aggregating modern pixel embeddings into patch-level representations.

In this paper, we study the post-hoc pooling step in the fixed-embedding setting.
Given pixel embeddings from a released geospatial embedding product, how should a practitioner aggregate them into a patch-level representation when the encoder itself is unavailable?
We benchmark 11 training-free methods and 2 train-set-fitted baselines using pixel embeddings generated by three GFMs on the common EuroSAT land-cover classification dataset~\cite{helber2019eurosat}.
We evaluate the effects of pooling choice with kNN and linear probes under the standard random and geographically disjoint dataset splits.

Specifically, we release \emph{EuroSAT-Embed} and benchmark 13 pooling methods across three fixed embedding products, two probes, and both random and spatial splits.

\paragraph{Contributions.}
We make three contributions:
(1)~we release \emph{EuroSAT-Embed}, a set of three aligned pixel-level embedding datasets derived from AlphaEarth, OlmoEarth, and Tessera;
(2)~we provide a controlled benchmark of 11 training-free pooling methods and 2 train-set-fitted baselines across two probes and both random and spatial splits in the fixed-embedding setting;
(3)~we show that simple distributional pooling methods, especially \textit{stats}, consistently improve spatial generalization over mean pooling, while \textit{covariance} offers the strongest accuracy when higher-dimensional representations are acceptable.

\section{Data and Embeddings}

\textbf{EuroSAT.}
We use the 10-class EuroSAT land-cover dataset~\citep{helber2019eurosat} which contains 27,000 Sentinel-2 patches at 64$\times$64 pixels with classes such as forest, residential, and agricultural land.
We evaluate on both the standard random split~\citep{neumann2019domain} and a spatial split~\citep{ekim2025distribution} that partitions samples by longitude to reduce train-test leakage via spatial autocorrelation.

\textbf{EuroSAT-Embed.}
We construct three aligned embedding datasets from AlphaEarth~\citep{brown2025alphaearth} (64-d), OlmoEarth-Nano~\citep{herzog2025olmoearth} (128-d), and Tessera~\citep{feng2025tesseratemporalembeddingssurface} (128-d).
AEF and OlmoEarth embed patch context but emit \emph{pixel-wise} vectors, whereas Tessera encodes per-pixel time-series without spatial context of adjacent pixels.
We use these embedding products to study pooling behavior, not to compare the underlying GFMs.

\section{Pooling Methods}

We benchmark training-free operators that compress an $H{\times}W{\times}D$ embedding tensor into a $d$-dimensional vector.
We hypothesize that class signal lies not only in the mean but in the distribution (variance and extremes).
For example, a residential neighborhood might share a mean embedding over space with industrial areas, but show higher variance from mixed rooftops, vegetation, and roads.
Let $X \in \mathbb{R}^{H \times W \times D}$ denote a patch with $N{=}HW$ pixels.
Pooling operates per channel (e.g., $\max(X)$ returns the per-channel maximum). Note that pooling methods can increase the dimensionality by a multiplier due to concatenation of multiple statistics.

\textbf{First-order statistics} ($D$-d each):
\textit{mean} ($\mu = \frac{1}{N}\sum_i x_i$) captures the ``typical'' pixel but
discards variation; \textit{max} preserves extreme activations; generalized mean
(\textit{GeM}) interpolates between them~\citep{radenovic2018gem};
\textit{center-weighted mean} down-weights boundary pixels.

\textbf{Distributional statistics} ($D$-d to $5D$-d):
\textit{std} captures pixel variability; \textit{mean+std} ($2D$-d)
and \textit{mean+max} ($2D$-d)
concatenate complementary signals; \textit{stats} ($4D$-d) stacks
min/max/mean/std;
\textit{percentiles} ($5D$-d) computes five summary quantiles;
\textit{median+IQR} ($2D$-d) provides robust location and spread estimates;
\textit{covariance} extracts the upper triangle of the pixel covariance matrix
($D(D{+}1)/2$-d).

\textbf{Parametric methods} (fit after embeddings are produced):
\textit{PCA} projects mean-pooled embeddings to lower dimensions;
\textit{Bag of Visual Words (BoVW)}~\citep{csurka2004visual} clusters
pixel embeddings with mini-batch $k$-means and
represents each patch as a normalized histogram of cluster assignments.
These still operate in the post-hoc setting, but unlike the training-free
pooling operators above, they fit an additional summarization stage on the
training split.

\begin{table*}[t]
  \centering
  \footnotesize
  \caption{Linear probe accuracy across embedding sources. Cell shading indicates higher accuracy within each column (darker = higher). Best per column in \textbf{bold} and second-best in \textit{italics}. Gap = accuracy drop from random to spatial split. $^*$OlmoEarth-Nano variant.}\label{tab:linear_all}
  \begin{tabular}{lcccc|cccc|c}
    \toprule
    & \multicolumn{4}{c}{Spatial split} & \multicolumn{4}{c}{Random split} & \\
    \cmidrule(lr){2-5} \cmidrule(lr){6-9}
    Pooling & AEF & OlmoEarth$^*$ & Tessera & Avg & AEF & OlmoEarth$^*$ & Tessera & Avg & Gap$\downarrow$ \\
    \midrule
    Std & \cellperf{15}{78.6} & \cellperf{33}{90.6} & \cellperf{40}{87.8} & \cellperf{24}{85.7} & \cellperf{15}{90.7} & \cellperf{33}{94.8} & \cellperf{34}{95.5} & \cellperf{20}{93.6} & \cellperf{36}{8.0} \\
    Mean & \cellperf{34}{84.0} & \cellperf{46}{92.3} & \cellperf{33}{85.5} & \cellperf{32}{87.3} & \cellperf{53}{95.5} & \cellperf{51}{96.5} & \cellperf{41}{96.1} & \cellperf{47}{96.0} & \cellperf{31}{8.8} \\
    GeM & \cellperf{38}{85.0} & \cellperf{39}{91.4} & \cellperf{37}{86.9} & \cellperf{34}{87.8} & \cellperf{54}{95.6} & \cellperf{45}{95.9} & \cellperf{46}{96.5} & \cellperf{47}{96.0} & \cellperf{34}{8.2} \\
    Center-Weighted & \cellperf{38}{84.9} & \cellperf{47}{92.4} & \cellperf{38}{87.0} & \cellperf{36}{88.1} & \cellperf{58}{96.2} & \cellperf{53}{96.7} & \cellperf{43}{96.3} & \cellperf{52}{96.4} & \cellperf{34}{8.3} \\
    Max & \cellperf{42}{86.0} & \cellperf{36}{91.1} & \cellperf{46}{89.7} & \cellperf{40}{88.9} & \cellperf{32}{92.9} & \cellperf{15}{93.0} & \cellperf{15}{93.8} & \cellperf{15}{93.2} & \cellperf{57}{4.3} \\
    Median+IQR & \cellperf{33}{83.6} & \cellperf{53}{93.3} & \cellperf{52}{91.7} & \cellperf{43}{89.5} & \cellperf{46}{94.6} & \cellperf{47}{96.1} & \cellperf{42}{96.2} & \cellperf{43}{95.6} & \cellperf{47}{6.1} \\
    Mean+Std & \cellperf{29}{82.4} & \cellperf{49}{92.7} & \cellperf{58}{93.9} & \cellperf{44}{89.7} & \cellperf{57}{96.0} & \cellperf{52}{96.5} & \cellperf{55}{97.3} & \cellperf{54}{96.6} & \cellperf{42}{6.9} \\
    Percentiles & \cellperf{47}{87.4} & \cellperf{57}{\textit{93.8}} & \cellperf{55}{92.8} & \cellperf{52}{91.3} & \cellperf{57}{96.0} & \cellperf{54}{\textit{96.8}} & \cellperf{53}{97.1} & \cellperf{55}{96.6} & \cellperf{51}{5.3} \\
    Mean+Max & \cellperf{50}{88.2} & \cellperf{53}{93.2} & \cellperf{58}{93.9} & \cellperf{54}{91.8} & \cellperf{59}{\textit{96.3}} & \cellperf{48}{96.1} & \cellperf{49}{96.7} & \cellperf{52}{96.4} & \cellperf{56}{4.6} \\
    Covariance & \cellperf{60}{\textbf{91.0}} & \cellperf{51}{92.9} & \cellperf{60}{\textbf{94.4}} & \cellperf{59}{\textit{92.8}} & \cellperf{59}{96.2} & \cellperf{60}{\textbf{97.3}} & \cellperf{60}{\textbf{97.7}} & \cellperf{60}{\textbf{97.1}} & \cellperf{57}{\textit{4.3}} \\
    Stats & \cellperf{59}{\textit{90.6}} & \cellperf{60}{\textbf{94.1}} & \cellperf{59}{\textit{94.3}} & \cellperf{60}{\textbf{93.0}} & \cellperf{60}{\textbf{96.4}} & \cellperf{54}{96.8} & \cellperf{56}{\textit{97.4}} & \cellperf{57}{\textit{96.8}} & \cellperf{60}{\textbf{3.8}} \\
    \midrule
    \multicolumn{10}{l}{\textit{Parametric pools (fit on train set)}} \\
    PCA & \cellperf{23}{80.8} & \cellperf{39}{91.5} & \cellperf{15}{79.3} & \cellperf{15}{83.9} & \cellperf{52}{95.4} & \cellperf{45}{95.9} & \cellperf{29}{95.0} & \cellperf{41}{95.4} & \cellperf{15}{11.6} \\
    BoVW & \cellperf{53}{89.0} & \cellperf{15}{88.3} & \cellperf{43}{88.7} & \cellperf{39}{88.7} & \cellperf{46}{94.6} & \cellperf{29}{94.4} & \cellperf{46}{96.5} & \cellperf{37}{95.1} & \cellperf{45}{6.5} \\
    \bottomrule
  \end{tabular}
\end{table*}

\section{Experimental Setup}

We evaluate two probes:
kNN ($k{=}5$, cosine distance) and multinomial logistic regression, selecting
regularization $C$ for each pooling method by 3-fold cross-validation on the
training split only.
kNN probes geometry; linear probes test linear separability.
We standardize before linear probing.
We evaluate on both random and geographically disjoint (spatial) splits.
Results are reported in Tables~\ref{tab:linear_all},~\ref{tab:knn_all}.

\section{Results}

\begin{wrapfigure}{r}{0.48\linewidth}
  \vspace{-12pt}
  \centering
  \includegraphics[width=\linewidth]{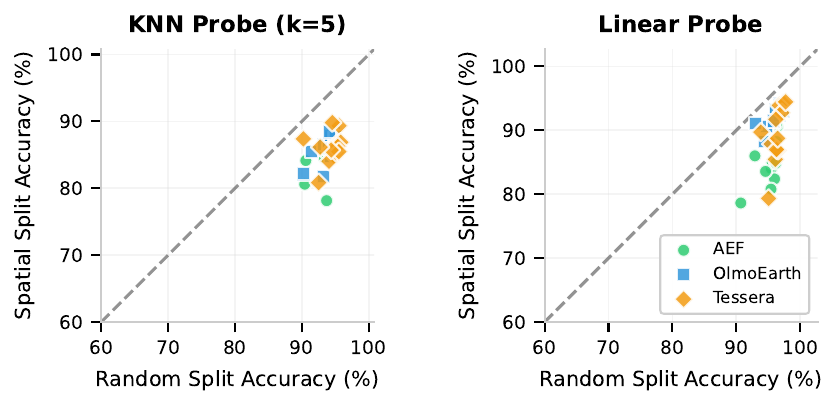}
  \caption{\textbf{Random vs spatial accuracy.} Points near the
    diagonal (where random $=$ spatial) have smaller generalization gaps.
  }\label{fig:random_spatial}
  \vspace{-8pt}
\end{wrapfigure}

\textbf{Distributional statistics improve generalization.}
For linear probes, \textit{stats} pooling reaches 91--94\% on spatial splits
(avg 93.0\%) versus 84--92\% for \textit{mean} (avg 87.3\%), a +6\% gain
(Table~\ref{tab:linear_all}).
\textit{Covariance} pooling achieves the highest accuracy on two of three
encoders (AEF: 91.0\%, Tessera: 94.4\%) at $D(D{+}1)/2$ dimensions.
On random splits, differences narrow to ${\sim}$1\% as spatial leakage
boosts all methods.

kNN results (Table~\ref{tab:knn_all}) show consistent trends:
\textit{stats} leads on spatial splits (avg 88.0\%) while
\textit{center-weighted mean} leads on random splits (avg 95.3\%).
Second-order statistics (\textit{covariance}) help kNN less
than linear probes, suggesting that they benefit linear classification more
than similarity search.

\textbf{Generalization gaps.}
Figure~\ref{fig:random_spatial} shows methods diverge under distribution shift.
\textit{Mean} pooling drops 8.8\,pp (random to spatial); \textit{stats} drops
only 3.8\,pp---a 57\% reduction in gap.
\textit{Covariance} (4.3\,pp) and \textit{mean+max} (4.6\,pp)
also exhibit small gaps while maintaining high accuracy.

\begin{table*}[t]
  \centering
  \footnotesize
  \caption{KNN accuracy ($k{=}5$) across embedding sources. Cell shading indicates higher accuracy within each column (darker = higher). Best per column in \textbf{bold} and second-best in \textit{italics}. Gap = accuracy drop from random to spatial split. $^*$OlmoEarth-Nano variant.}\label{tab:knn_all}
  \begin{tabular}{lcccc|cccc|c}
    \toprule
    & \multicolumn{4}{c}{Spatial split} & \multicolumn{4}{c}{Random split} & \\
    \cmidrule(lr){2-5} \cmidrule(lr){6-9}
    Pooling & AEF & OlmoEarth$^*$ & Tessera & Avg & AEF & OlmoEarth$^*$ & Tessera & Avg & Gap$\downarrow$ \\
    \midrule
    Std & \cellperf{15}{78.1} & \cellperf{15}{81.7} & \cellperf{31}{84.0} & \cellperf{15}{81.3} & \cellperf{45}{93.7} & \cellperf{44}{93.3} & \cellperf{46}{94.1} & \cellperf{45}{93.7} & \cellperf{15}{12.4} \\
    Max & \cellperf{28}{80.6} & \cellperf{41}{85.5} & \cellperf{15}{80.8} & \cellperf{22}{82.3} & \cellperf{15}{90.4} & \cellperf{26}{91.4} & \cellperf{33}{92.5} & \cellperf{25}{91.4} & \cellperf{37}{9.1} \\
    GeM & \cellperf{54}{85.8} & \cellperf{47}{86.6} & \cellperf{38}{85.4} & \cellperf{46}{85.9} & \cellperf{53}{94.6} & \cellperf{44}{93.3} & \cellperf{54}{95.1} & \cellperf{51}{94.3} & \cellperf{42}{8.4} \\
    Median+IQR & \cellperf{56}{86.1} & \cellperf{44}{86.0} & \cellperf{39}{85.7} & \cellperf{46}{85.9} & \cellperf{48}{94.1} & \cellperf{50}{94.0} & \cellperf{48}{94.4} & \cellperf{49}{94.1} & \cellperf{44}{8.2} \\
    Mean & \cellperf{53}{85.6} & \cellperf{56}{87.8} & \cellperf{35}{84.7} & \cellperf{47}{86.0} & \cellperf{53}{94.6} & \cellperf{58}{94.8} & \cellperf{53}{94.9} & \cellperf{55}{94.8} & \cellperf{40}{8.7} \\
    Covariance & \cellperf{47}{84.5} & \cellperf{36}{84.8} & \cellperf{60}{\textbf{89.8}} & \cellperf{49}{86.4} & \cellperf{42}{93.4} & \cellperf{53}{94.3} & \cellperf{50}{94.5} & \cellperf{49}{94.1} & \cellperf{47}{7.7} \\
    Center-Weighted & \cellperf{56}{86.1} & \cellperf{56}{87.8} & \cellperf{38}{85.5} & \cellperf{50}{86.5} & \cellperf{60}{\textbf{95.4}} & \cellperf{60}{\textbf{95.0}} & \cellperf{58}{95.5} & \cellperf{60}{\textbf{95.3}} & \cellperf{39}{8.8} \\
    Mean+Max & \cellperf{52}{85.4} & \cellperf{56}{87.9} & \cellperf{45}{86.9} & \cellperf{52}{86.7} & \cellperf{54}{94.7} & \cellperf{54}{94.4} & \cellperf{60}{\textbf{95.8}} & \cellperf{57}{94.9} & \cellperf{43}{8.2} \\
    Percentiles & \cellperf{59}{\textit{86.8}} & \cellperf{54}{87.5} & \cellperf{42}{86.2} & \cellperf{52}{86.8} & \cellperf{53}{94.6} & \cellperf{59}{94.9} & \cellperf{56}{95.4} & \cellperf{57}{95.0} & \cellperf{44}{8.1} \\
    Mean+Std & \cellperf{58}{86.5} & \cellperf{59}{\textit{88.3}} & \cellperf{44}{86.7} & \cellperf{55}{\textit{87.2}} & \cellperf{55}{94.9} & \cellperf{59}{\textit{94.9}} & \cellperf{56}{95.3} & \cellperf{58}{\textit{95.0}} & \cellperf{46}{7.9} \\
    Stats & \cellperf{60}{\textbf{87.0}} & \cellperf{54}{87.6} & \cellperf{58}{\textit{89.3}} & \cellperf{60}{\textbf{88.0}} & \cellperf{57}{\textit{95.0}} & \cellperf{54}{94.4} & \cellperf{58}{\textit{95.6}} & \cellperf{57}{95.0} & \cellperf{52}{7.0} \\
    \midrule
    \multicolumn{10}{l}{\textit{Parametric pools (fit on train set)}} \\
    BoVW & \cellperf{46}{84.1} & \cellperf{18}{82.1} & \cellperf{48}{87.4} & \cellperf{37}{84.6} & \cellperf{17}{90.6} & \cellperf{15}{90.3} & \cellperf{15}{90.2} & \cellperf{15}{90.4} & \cellperf{60}{\textbf{5.8}} \\
    PCA & \cellperf{52}{85.4} & \cellperf{60}{\textbf{88.4}} & \cellperf{42}{86.1} & \cellperf{51}{86.7} & \cellperf{43}{93.5} & \cellperf{52}{94.1} & \cellperf{35}{92.7} & \cellperf{43}{93.5} & \cellperf{53}{\textit{6.8}} \\
    \bottomrule
  \end{tabular}
\end{table*}

\section{Discussion}

\textbf{Recommendation.}
For practitioners working with fixed pixel-level embedding products, pooling
should be treated as a first-order design choice rather than a default
preprocessing step.
\textit{Mean} pooling ($1\times$) is a useful baseline but leaves significant
accuracy on the table under spatial shift.
When a modest increase in representation size is acceptable, \textit{stats}
pooling ($4\times$) is the strongest overall default across encoders and probes
on spatial splits.
When representation size is less constrained and maximum accuracy is the
priority, \textit{covariance} pooling ($D(D{+}1)/2$-d) is often strongest.

\textbf{Accuracy and robustness are aligned.}
We do not observe an accuracy--robustness tradeoff: methods with the highest
spatial-split accuracy also tend to have the smallest random-to-spatial
generalization gaps. In particular, \textit{stats}, \textit{covariance}, and
\textit{mean+max} are both more accurate and more robust than \textit{mean}.

\textbf{Why distributional statistics help.}
\textit{Mean} pooling collapses each patch to a first-moment summary, discarding intra-patch heterogeneity; distributional pooling preserves variability that appears useful under spatial shift.

\textbf{Limitations.}
We evaluate on a single benchmark (EuroSAT, 10 classes) with three
embedding sources;
broader coverage across datasets such as BigEarthNet and fMoW, and multi-label settings,
would strengthen conclusions.
We focus on training-free pooling; learned methods such as attention pooling
or adaptive GeM may improve further but require task-specific
training~\citep{touvron2022augmenting}.
Our conclusions should be interpreted as guidance for post-hoc pooling of fixed
embedding products rather than for end-to-end learned aggregation with encoder
access.
On EuroSAT-Embed, pooling choice materially changes spatial-split
performance, and simple distributional summaries are a strong default.

\bibliography{iclr2026_conference}
\bibliographystyle{iclr2026_conference}

\end{document}